\documentclass[10pt,twocolumn,letterpaper]{article}

\usepackage{cvpr}
\usepackage{times}
\usepackage{epsfig}
\usepackage{float}
\usepackage{graphicx}
\usepackage{subcaption}
\usepackage{enumerate}
\usepackage{amsmath}
\usepackage{amssymb}

\usepackage[pagebackref=true,breaklinks=true,letterpaper=true,colorlinks,bookmarks=false]{hyperref}

\cvprfinalcopy % *** Uncomment this line for the final submission

\def\cvprPaperID{9717} % *** Enter the CVPR Paper ID here

\ifcvprfinal\fi
\begin{document}

%%%%%%%%% TITLE
\title{ Direct Classification of Emotional Intensity}

\author{Jacob Ouyang\\
Unversity of California Irvine, AiCure\\
Irvine, California\\
{\tt\small jnouyang@uci.edu}
\and
Isaac Galatzer-Levy\\
AiCure\\
New York City, New York\\
{\tt\small Isaac.Galatzer-Levy@aicure.com}
\and
Vidya Koesmahargyo\\
AiCure\\
New York City, New York\\
{\tt\small vidya.koesmahargyo@aicure.com}
\and
Li Zhang\\
AiCure\\
New York City, New York\\
{\tt\small li.zhang@aicure.com}
}
\maketitle
%\thispagestyle{empty}

%%%%%%%%% ABSTRACT
\begin{abstract}
In recent years, computer vision models have gotten significantly better at understanding emotion. However, there has been limited research in predicting the emotional intensity in people's expressions which is necessary in a variety of applications from disease detection to driver drowsiness. Current models use action units and derive emotional intensities based on action units. In this paper, we present a model that can directly predict emotion intensity score from video inputs, instead of deriving from action units. Using a 3d DNN incorporated with dynamic emotion information, we train a model using videos of different people smiling that outputs an intensity score from 0-10. Each video is labeled framewise using a normalized action-unit based intensity score. Our model then employs an adaptive learning technique to improve performance when dealing with new subjects. Compared to other models, our model excels in generalization between different people as well as provides a new framework to directly classify emotional intensity.
\end{abstract}

%%%%%%%%% BODY TEXT
\section{Introduction}
Automatic recognition of spontaneous emotion have significant impact on human-computer interactions, as well as in emotion-related studies in fields such as psychiatry, where emotion intensity detection can play a vital role. Emotion intensity recognition has been used for a variety of applications, including identifying clinical populations such as Autism Spectrum Disorder and Schizophrenia \cite{monkul_green_barrett_robinson_velligan_glahn_2007,Griffiths2019}, detection of a drowsy driver \cite{vural2008automated}, and studying differential patterns of emotion perception in various diseases \cite{10.3389/fpsyg.2012.00098}. Moreover, automated detection of emotional intensity has many potential applications to not only aid in clinical identification of a disease, but also in monitoring its progression. For example, automated emotional intensity detection could be used in remotely assessing for patients without added burden to the clinician or patient.

	Certain facial muscle activations, known as FACS (Facial Action Coding System) action units, have been linked to emotional expressivity.\cite{CK+} Many attempts have been made to automatically label action units, and their intensities are used to classify emotions. \cite{WildDeepNets, EmotioNet, DeepCoder, OpenFace, PartiallyLabeled}. While these have achieved state of the art success, these models rely solely on action units, which represent only a subset of facial markers. Other attempts have been made to create a larger feature space that contains more facial markers \cite{Chang1}, although these are much harder to train. Emotion can also be directly classified from facial images \cite{Tautkute1}. These models only target the presence of an emotion, rather than the expressivity intensity, which alone does not give enough information in clinical applications such as identifying autism.

	Due to the variance in expressivity of emotions, there has not been a standardized method for labeling emotional intensity. Even if relative frequencies of emotions can be determined by comparison, there has been no attempt to create a standardized cross population labeling scheme for emotional intensity.

In our paper, we tackle four primary challenges of creating an emotional intensity model.
\begin{enumerate}[1)]
\item
There is no standardized system that measures emotional intensity. Even though action units give intensities that are correlated to the presence of an emotion, there is no standardized mapping from AU to emotional intensity.
\item
The majority of the computer vision approaches related to emotion / AU are static image based, while our method quantifies emotion from the dynamic information in videos.
\item
There are differences in how different people express varied levels of emotion. This makes it hard to establish a benchmark for different intensities across different subjects.
\item
Lack of balanced emotion data homogeneously distributed on various intensity labels makes training difficult.
\end{enumerate}

%-------------------------------------------------------------------------
\section{Methods}
%-------------------------------------------------------------------------
\subsection{Emotion Labeling Scheme}

	Researchers have long linked action units to emotions. Happiness, for example, have been linked to Action Units 6 and 12, Orbicularis oculi and Zygomatic Major respectively. Other emotions and their respective action units are shown in Figure 1.
\begin{figure}[h!]
\centering
\includegraphics[width=0.8\columnwidth,scale=0.8]{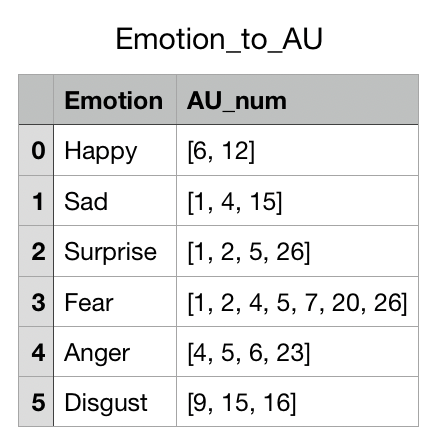}
\caption{Emotion to Action Unit Mapping}
\end{figure}
	Action units provide a good metric to detect the presence of emotions. However, there has been very limited research in the direct correlation between action unit intensity and the amount of emotional expressivity between various patients. The maximum expressivity of an action unit for a given expression may also be highly variant. For example, in happiness, AU 6 varies between 0.4 and 0.6 at max expression. Thus, it’s hard to create a standardized correlation between action unit and emotional intensity. As such, we create a normalized set of action units to use for our emotional score.
For each patient in our training dataset, we assume that we receive their full range of expression in the video. Thus, each action unit should have reached a maxima for that emotion during their max expressivity. We use these constants to create our label. For happiness, our labeling scheme for a given frame would be such:

\begin{align*}
Happiness_{i}&=\left(\frac{AU_{6Li}+AU_{6Ri}}{max(AU_{6L})+max(AU_{6R})}\right)/2\\
&+\left(\frac{AU_{12Li}+AU_{12Ri}}{max(AU_{12L})+max(AU_{12R})}\right)/2
\tag{1}
\end{align*}
In general, our equation for a given emotion based on its action units would be:
\begin{align*}
Emotion_{x}=\sum_{i=1}^{n}\frac{AU_{xi}}{max(AU_{xi})}*\frac{1}{n}\\
\tag{2}
\end{align*}
For n action units that make up Emotion X.
%------------------------------------------------------------------------
\begin{figure*}[htbp!]
\centering
\includegraphics[width=.8\textwidth, height=7cm]{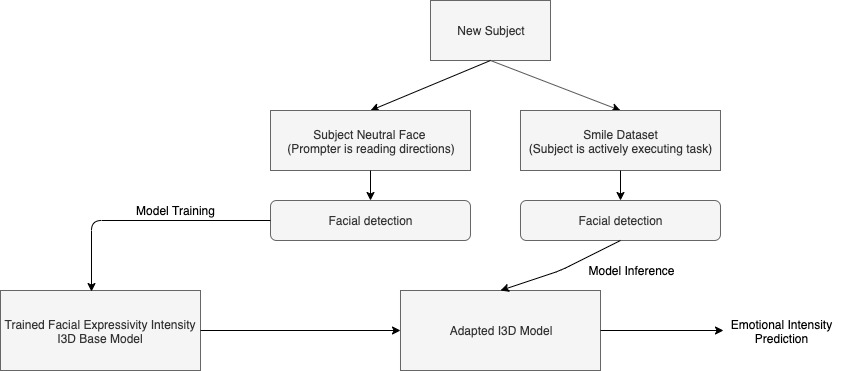}
\caption{New subject goes through Inference Pipeline}
\end{figure*}
\subsection{Video Based Approach}

	A fundamental difference between our approach and those using action units is our model’s ability to analyze several frames, which would contain information about motion and time. Action units only look at the amount of muscle activation (the offset between one's neutral state and their current), but with a multitude of frames, the model can look for change in facial color and movement. We use the Inception 3D RGB network \cite{I3D} as our base model. This model is an inflated-version of the 2D Inception network that opts to use 3D convolutions with dimensions $N\times N \times N$ to replace 2D conv layers that have $N\times N$ dimensions. The dataset has been pre trained with ImageNet as well the Kinetics Dataset, so it excels at action recognition and is a prime candidate for transfer learning.

For the purpose of our model, we only trained the model on happiness, as we found it to be the most easily reproducible emotion and its action units most reflective of expression.

%------------------------------------------------------------------------
\subsection{Data Augmentation with Transitional States}
One problem with learning directly off the videos in our dataset was how unbalanced our video  labels were. Lower intensity expressions (normalized labels between 3 and 7), occupy $\frac{1}{4}$ of the total dataset. Thus, models would be unable to distinguish expression at different intensities and labels were pushed towards extremes.
 	To solve this, we introduce a data augmentation method to introduce variation in video data. Because intensity rapidly increases during changes in expressivity (from neutral to full expressivity), we can augment frames when the video is in a transitional state, effectively creating videos which show subjects expressing lower intensities. This has shown to be an effective method in not only balancing classes, but in generating "new" data.
%------------------------------------------------------------------------
\subsection{Adaptive Learning}
Facial models are often able to learn on known faces but have been known to be weaker at generalization to new faces. To solve this problem, we employ a technique we call “adaptive learning.” Given a new subject used for model inference, we divide their data into two subsets: neutral face and active expression. For a given task, we assume that the subject is able to maintain a neutral state, in which they are not actively exerting a facial expression. We call this subset the neutral face dataset. This dataset is then used to finetune the I3D model trained with our training datasets. The fine tuning follows the same procedure as the training dataset.

Figure 2 shows is a graphical interpretation of the adaptive learning process. Neutral faces are used to train the model, whereas the smile dataset is used for inference.

%------------------------------------------------------------------------
\section{Implementation}
\subsection{Preprocessing}
	Our model employs some classical computer vision preprocessing techniques as well as some emotion specific data augmentation methods that we will detail in this section. As for standard preprocessing techniques, we first ran OpenCV’s \cite{OpenCV} DL facial detection model on the video. We created a wide bounding box, using the largest width and length created over a video segment to minimize noise from head movement as well as background noise. We also use a variety of standard data augmentation techniques when training the model. We use both random cropping and brightness adjustment to allow for more noisy training environments and prevent overfitting. We then crop each image to 224x224 when feeding into the model.

%------------------------------------------------------------------------
\subsection{Model Input}
\begin{figure*}[htb!]
\centering
\includegraphics[width=\textwidth, height=8cm]{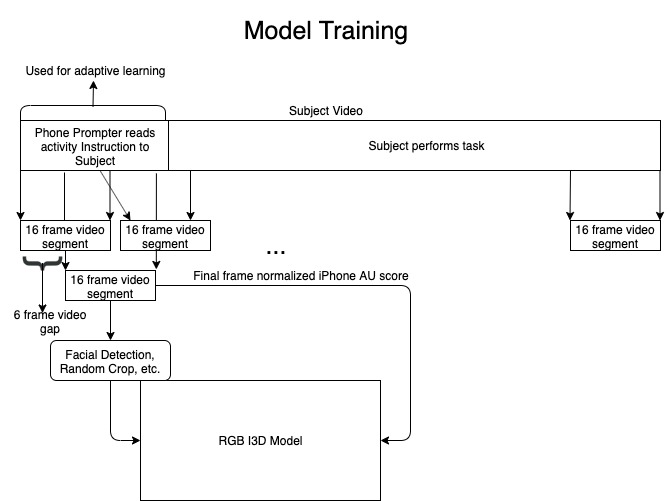}
\caption{Video Based Model Input for RGB I3D Network}
\end{figure*}
	For each frame, we used the last 16 frames of video up to that frame as input to the network, around ~0.5 seconds. This time frame allows the model to recognize movement in the model, while also not containing too much noise. The average subject in our training set was able to go from neutral face to a full smile in about 0.5 seconds. We used the action unit label of the most recent frame in the 16 frame input as our label for the sequence. Based on the normalized AU score we created above, we rounded it to the nearest tenth, and assign it that number as a label.
%------------------------------------------------------------------------
\subsection{Modified I3D Model}
	As stated before, we use the Inception 3d RGB Network (I3D RGB)\cite{I3D} as our base network and we use checkpoints based on initial training from inflated Imagenet and Kinetics Dataset. We tried both regression and classification models for this network and found that the classification model outperforms the regression model. As such, we modified the last 1x1x1 3D convolution layer from having 1024 output nodes to 11, with randomly initialized weights mapping intensity 0,0.1,0.2…,1.0 to the respective nodes 0,1,2,..,10. We use cross entropy loss and standard gradient descent to train our model.
%------------------------------------------------------------------------
\subsection{HyperParameters}
	We split the training set by subjects with a 70/30 training validation split. This resulted in 26/15 patient split. We use 16 frames as the input into the model, so we create  16 frame video segments which start 6 frames away from the previous video. This results in a 10 frame overlap, and gives us a grand total of 3552 and 1998 videos segments respectively for training and validation sets. We apply the preprocessing techniques highlighted above. The network uses a learning rate of 0.0001, momentum 0.9, a scheduler with alpha 0.1 every 7 epochs. We ran 30 epochs of our training set.
%------------------------------------------------------------------------
\section{Results}
%------------------------------------------------------------------------
\subsection{Dataset}
	Our dataset included 41 subjects, with each subject having several 3-10 second videos of them smiling. Each subject starts with a neutral face, and then we asked each of the subjects to smile, followed by smile more, creating videos of them smiling while peeking at different intensities. For each subject, we assume that they are capable of showing their max range of expression as a result of our multi step process.

	These videos were all taken on iPhoneX, using their action unit labels to create our ground truth. iPhoneX data is known to include both RGB and depth data, so we assume that its labeling will use a larger amount of information. For each video, the iPhone returned a list of Action Unit intensities for every frame where a face was detected.
\begin{figure*}[htbp!]
\centering
\includegraphics[width=\textwidth, height=6cm]{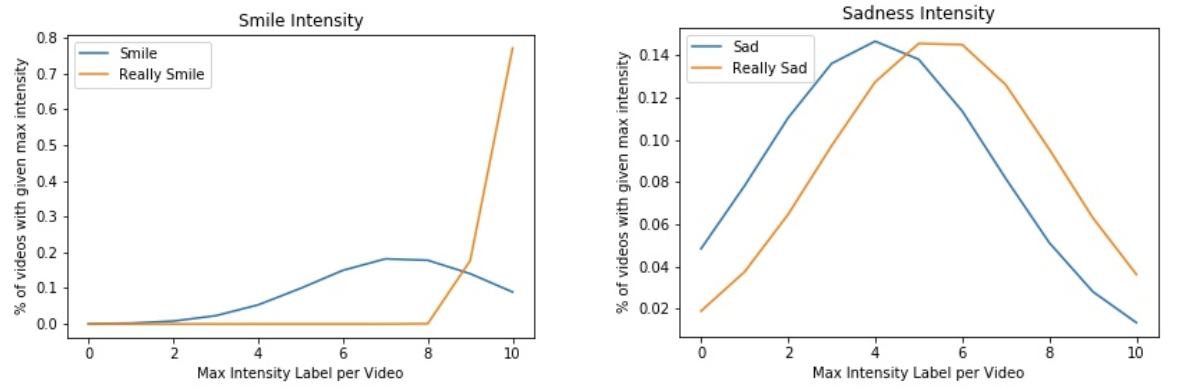}
\caption{Max Intensity Comparison between Normal and Max Expression}
\end{figure*}

	The 41 subjects were collected in two main batches, each with its own different lighting and background. The first dataset, Dataset 1, was collected in a phone booth setting, with each subject having 2 videos of approximately 9 seconds in length. The subjects were asked to perform tasks where they were asked smile then really smile, be sad then really sad, angry, really angry etc. In these videos, subjects started with neutral expression, exhibited peak expression halfway through the video, and returned to neutral expression at the end.

	The second dataset, Dataset 2, was collected in a noisy environment with bright lights, with each subject having around 3 videos, each approximately 4 seconds in length. The subjects were asked to smile at 3 different intensities, be sad at 3 different intensities, etc. In this dataset, subjects were recorded only when they were exhibiting emotion.

	We primarily use DataSet for 2 for our training set and Dataset 1 for our validation set as it is easier to validate our model's performance on longer videos with a more clean expressivity curve.

	Out of the ~5500 videos in Datasets 1 and 2, roughly 1500 videos are created as a result of using our data augmentation with transition states method. This gives a roughly balanced dataset across all intensity labels.

	We also have a third dataset of schizophrenic patients. These patients were all recorded on their own devices, and therefore without AU labels. Without ground truth labeling, this dataset can only be judged with anecdotal evidence, and thus is used as a testing dataset to understand the effects of our adaptive learning models. Furthermore, this dataset is used to validate clinical assumptions about emotion expression between healthy and schizophrenic patients. We will be testing our hypothesis that affected patient groups will be less able to show a full range of emotional expressivity.

%------------------------------------------------------------------------
\subsection{Efficacy of Emotion Labeling}

	Figure 4 above show the difference between the max labels of different emotion labels in the task described in Dataset 1 given our labeling scheme described in Section 2.1. For happiness, there is a significant difference between expressivity in smiles versus really smile. This can be attributed to happiness being an easily artificially reproducible emotions with very clearly correlated action units. In our sadness labeling, there is still a separation between our tasks, even though the difference in the graphs is less evident. This behavior is consistent amongst the rest of the emotions (anger, disgust, surprise, fear) due to the other emotions being harder to artificially create, combined with action units that are less clearly correlated with expressivity.
	As such, we focus our results on smile, since we believe they have the most accurate labeling.
%------------------------------------------------------------------------
\subsection{Healthy Volunteer Data}
	After training for 30 epochs on the training set, the model with the highest accuracy yielded a 95\% perfect accuracy on the training set, but did not receive as optimal of results on the validation set. On validation set, it reached an accuracy of ~64\% within 2 labels of the original action unit labeling, and ~82\% within 3 labels. The base model can understand faces it’s seen well but fails to generalize on unknown faces. Furthermore, due to our small sample size of patients, it is likely overfitting on the training subjects. However, given our non-random validation accuracy, the model is still able to learn the underlying factors for happiness intensity.
	Our training set encompasses an incomplete set of ethnicities, backgrounds, lightings, and facial noise (ie. moustache). A majority of subjects in training are lighter skinned individuals with a dark background setting and clear lighting. Validation set patients that were of different or backgrounds or those with facial hair performed worse than the rest of the dataset. As such, the base model struggles more with generalization than overfitting. In the next section, we present our findings with adaptive learning that solve this problem.
%------------------------------------------------------------------------
\subsection{Adaptive Learning}
	One of the major struggles that emotion intensity neural network models face is their failure to generalize to new faces accurately. Without any further modification, it only had limited success on new faces. Moustaches, glasses, random facial movements (such as eating and tremors), all add noise into the model that can lead to inaccurate classification. Using a segment of the video where the patient is not exposed to any stimuli or executing a set of instructions. For the validation set, this set was the portion of the videos when the prompter were reading instructions.
\begin{figure}[h!]
\includegraphics[width=0.45\textwidth,height=4.5cm]{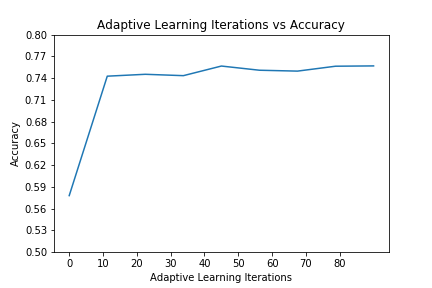}
\caption{Adaptive Learning Iteration vs Accuracy}
\end{figure}

This technique has been shown to significantly boost accuracy in the model’s ability to generalize intensity rankings. Figure 5 shows the average accuracy of intensity labeling across all subjects in the validation set over N iterations of neutral face. Each training iteration contains 1 sample of 3 videos of one subject taken from the neutral face dataset. Adaptive learning, on average, is able to boost model accuracy by 15\%.

\begin{figure}[h!]
\begin{subfigure}{.25\textwidth}
\includegraphics[width=\textwidth, height=5cm]{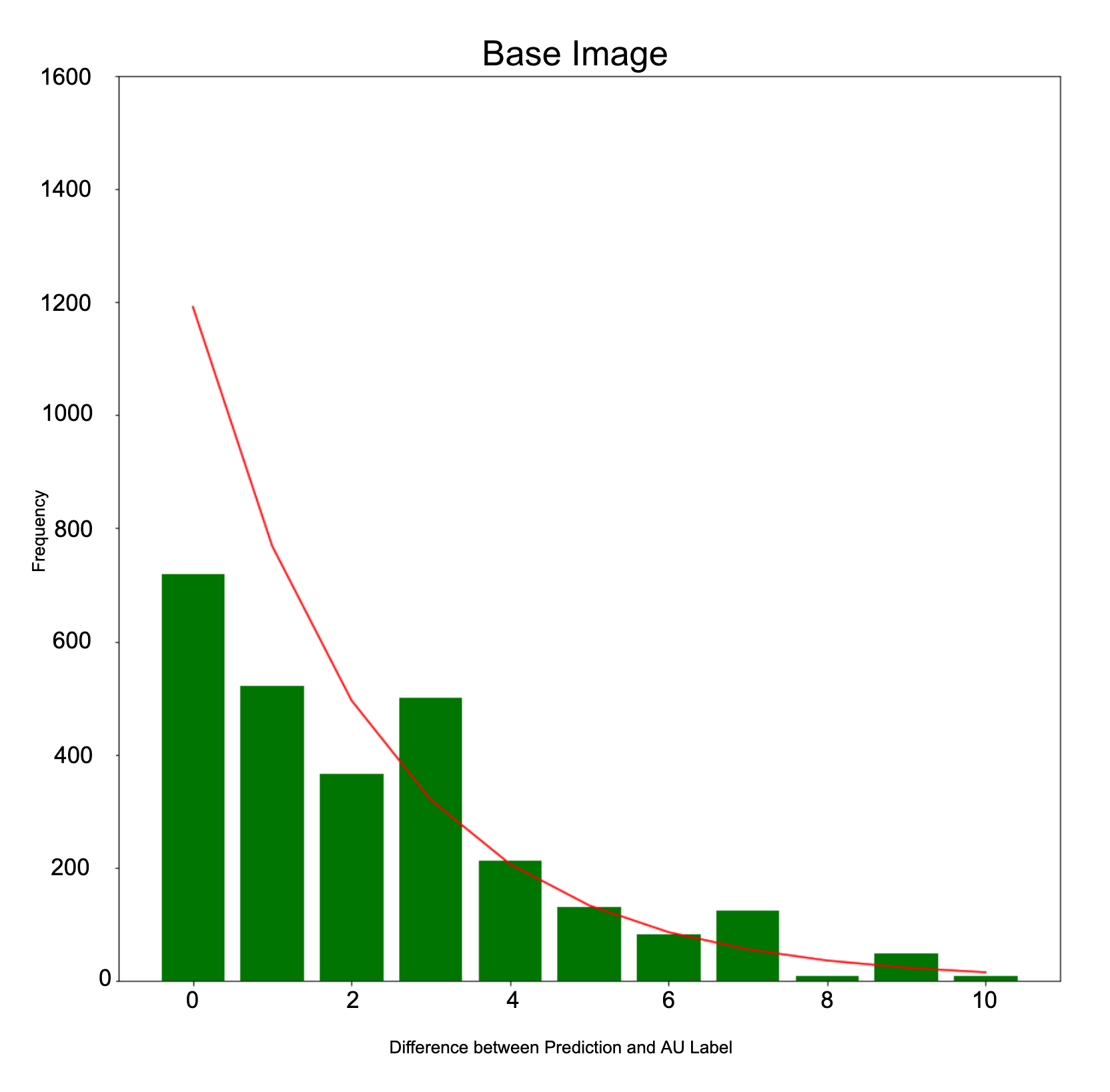}
\end{subfigure}%
\begin{subfigure}{.29\textwidth}
\includegraphics[width=0.9\textwidth,height=5cm]{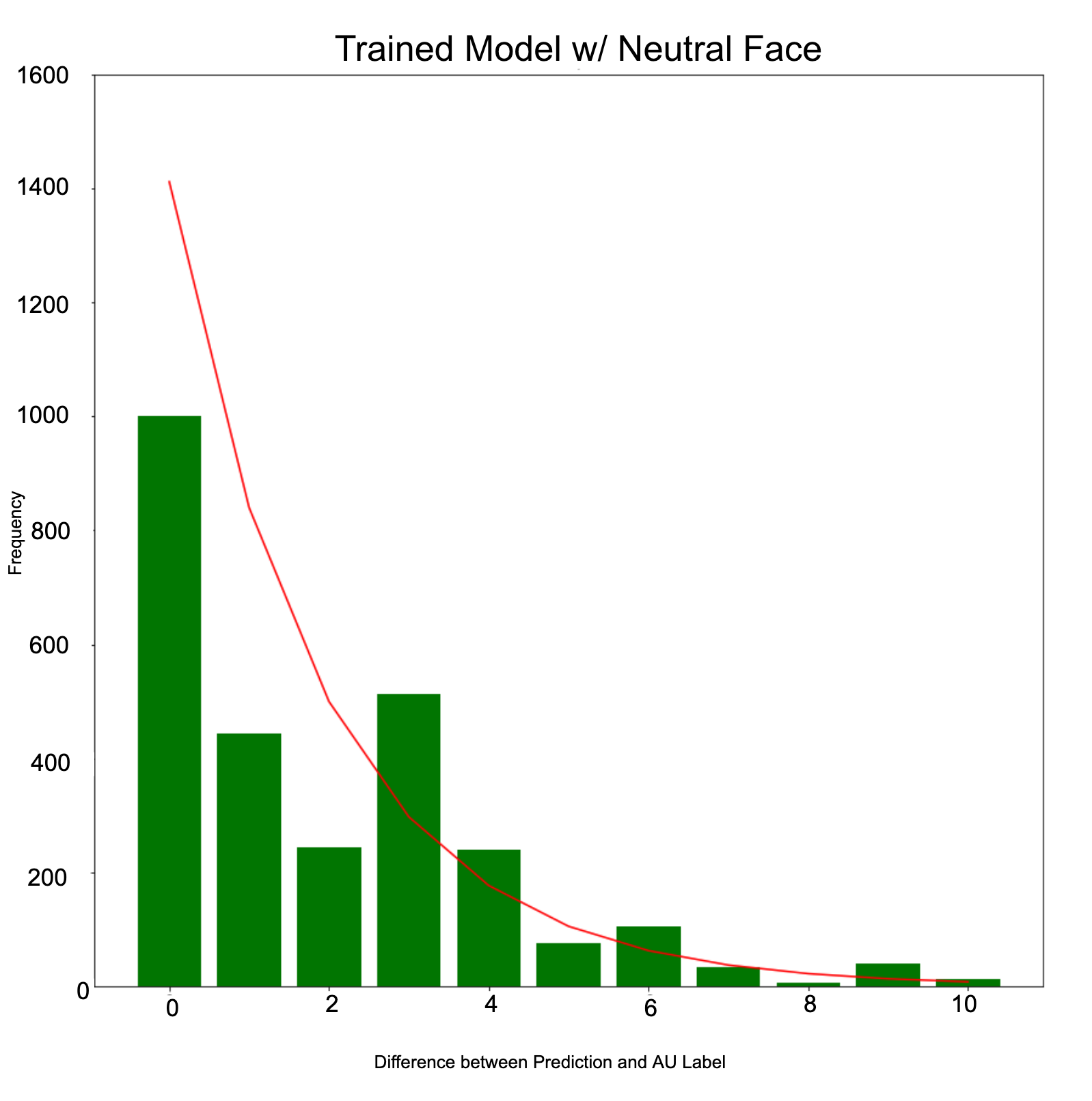}
\end{subfigure}
\caption{Difference in Label Accuracy before and after Adaptive Learning}
\end{figure}

	Figure 6 shows the difference between the original AU label and the model’s predicted label for each frame in the validation set. 0 would be a perfect prediction, 2 would be 2 labels off of the AU label, etc. This data does not include the video segments that were used for the adaptive learning dataset. The right figure, shows a clear improvement over the left, with an additional ~200 perfect labels. This neutral face method shows a significant increase in all new faces in the validation set.
%------------------------------------------------------------------------

\section{Discussions}
	The biggest issue with emotional intensity labeling is the lack of an accepted ground truth. In this paper, we introduce a new method of labeling intensity based on action units, which have been strongly linked with the presence of emotions. Our training dataset (Dataset 1 and 2) provide reliable AU labeling but still rely on the shaky assumption that a subject is able to provide a full range of expression to create our emotional intensity labels. In general, we assume that a difference of 0-2 labels between AU labeling and our model labeling is sufficient, to adjust for potential error in our normalization.

	With our patient dataset, we were unable to create ground truth labels. However, we still were able to use clustering to prove the effectiveness of our adaptive modeling. Our results with the schizophrenic dataset reflect our hypothesis that schizophrenic patients will react less to emotional stimuli when compared to healthy patients. More work is needed to visualize the results of this experiment

	Our adaptive learning technique is key in being able to reduce cross subject noise, as the approach effectively grounds neutral expression and forces the model to look for changes from base expression. Because we are looking for reaction to emotional stimuli in our prompts, we can assume that their neutral face has an expressivity score of 0.

\section{CONCLUSIONS \& FUTURE WORK}
This paper introduces	a new ground truth labeling system for emotional intensity. We have shown that the	 labeling scheme works well on happiness, but more research must be done in labeling other emotions (sadness, disgust, angry, surprise). Action unit correlations for non happiness emotions is not as direct, and forced facial expressions may not be as accurate for other emotions.

	Future work will be done to further our adaptive learning models. Using subjects with similar adapted model feature spaces, we can artificially train adaptive models on non neutral faces, using existing data to finetune adaptive models. Furthermore, more work can be done in using GANs to artificially generate emotions from neutral face using similar feature vectors and a given neutral faces. With this, we could potentially simulate expression in subjects without active data collection.
A significant amount of work is required in creating a dataset of natural expression that can be accurately labeled. Emotion intensity is a highly subjective field with no absolute ground truth, but our emotional rating still provides significant relative information. Clustering emotional intensity ratings can provide insight and potentially even alternate diagnosis patterns for affected patient groups such as those with schizophrenia.

{\small
\bibliographystyle{ieee_fullname}
\bibliography{egbib}
}

\end{document}